\documentclass{article}

\pdfoutput=1

\usepackage{amsmath}
\usepackage{microtype}
\usepackage{graphicx}
\usepackage{subcaption}
\usepackage{booktabs}
\usepackage{hyperref}

\usepackage{tikz}
\usetikzlibrary{arrows.meta, positioning, fit, calc, shapes.geometric}

\usepackage{algorithm}
\usepackage{algorithmic}

\usepackage[accepted]{icml2026}

\usepackage{amssymb}
\usepackage{mathtools}
\usepackage{amsthm}
\usepackage[capitalize,noabbrev]{cleveref}

\theoremstyle{definition}
\newtheorem{definition}{Definition}[section]
\newtheorem{problem}[definition]{Problem}
\theoremstyle{remark}

\usepackage[disable,textsize=tiny]{todonotes}

\icmltitlerunning{FactorLibrary}

\begin{document}

\twocolumn[
  \icmltitle{FactorLibrary: From Polynomials to Circuits via Recursive Subgoals}

  \icmlsetsymbol{equal}{*}

  \begin{icmlauthorlist}
  \icmlauthor{Rohan Pandey}{equal,uw}
  \icmlauthor{Michael Ruofan Zeng}{equal,uw}
  \icmlauthor{Weikun K. Zhang}{equal,uw}
  \icmlauthor{Kaijie Jin}{uw}
  \icmlauthor{Naomi Morato}{uw}
  \icmlauthor{Archit Ganapule}{uw}
  \icmlauthor{Bhaumik Mehta}{uw}
  \icmlauthor{Jarod Alper}{uw}
  \end{icmlauthorlist}

  \icmlaffiliation{uw}{University of Washington, Seattle, WA, USA}

  \icmlcorrespondingauthor{Michael Ruofan Zeng}{zengrf@uw.edu}

  \icmlkeywords{Arithmetic Circuits, Reinforcement Learning, Monte Carlo Tree Search, Symbolic Reasoning}

  \vskip 0.3in
]

\printAffiliationsAndNotice{\icmlEqualContribution}

\begin{abstract}
Finding minimal arithmetic circuits for polynomials over finite fields is a combinatorially hard problem central to algebraic complexity theory. We formulate it as a reinforcement learning problem in two directions, bottom-up and top-down. To address the challenge of a fast-growing combinatorial search space, we introduce \texttt{FactorLibrary}, which stores factorizable subexpressions that serve as reusable subgoals across training episodes. We trained a bottom-up agent with Gumbel-PPO-MCTS and two top-down agents with PPO+MCTS and SAC. The PPO+MCTS top-down agent exhibited the most stable performance, finding certified optimal circuits up to complexity $8$ with a success rate of \(91.8\%\). 
\end{abstract}

\section{Introduction}

An \textit{arithmetic circuit} is a sequence of addition and multiplication gates that computes a given polynomial from initial variable nodes \citep{Burgisser1997AlgebraicComplexity,ShpilkaYehudayoff2010ArithmeticCircuits}. A circuit is \textit{efficient} if it computes the polynomial with the minimal number of gates. The problem of finding efficient arithmetic circuits is central in algebraic complexity theory, and it also provides a useful testbed for reinforcement learning methods that search over a large combinatorial action space \citep{Valiant1979CompletenessClasses,Valiant1979Permanent,LidlNiederreiter1997FiniteFields}.  A much more difficult search problem of this sort is automatic generation of mathematical proofs, where learned policies must compose many verified symbolic steps \citep{Hubert2025AlphaProof}. 

In recent related work, \citet{Zhang2026CircuitBuilder} formulated the arithmetic circuit problem as a bottom-up reinforcement-learning game. Their ``\texttt{CircuitBuilder}'' agent begins with an empty circuit consisting of only variable and constant nodes, applies one gate at a time, and succeeds when a computed node equals the target polynomial. Zhang et al. compared PPO+MCTS with discrete SAC training architectures and observed that off-policy learning was competitive on easier two-variable tasks, while on-policy search performed better as the number of variables increased. 

In this paper, we adopt the \textit{bottom-up} approach of \citet{Zhang2026CircuitBuilder} and train with Gumbel-PPO-MCTS, where Gumbel search produces improved targets for PPO updates \citep{Danihelka2022GumbelPlanning,schulman2017ppo}. We also study an opposite \textit{top-down} formulation that reverses the direction of search. The agent starts from the target polynomial $f$ and repeatedly proposes addition split points $f = g+h$, relying on the game environment to produce lists of factors of $g$ and $h$. The agent then continues to recursively propose split points for smaller expressions until the search reaches the variable and constant nodes. The top-down formulation dramatically reduces the effective action space by exploiting deterministic factorization algorithms, and the results demonstrate this reduction clearly. 

Both formulations use a \texttt{FactorLibrary}, a collection of polynomials that serve as subgoals. It serves as both the agent's short-term memory and a list of recursive subgoals, under the heuristic that if the agent has learned to produce simpler subexpressions, then it should assemble them into the bigger expression. The technique of breaking down goals into recursive subgoals has been considered in \citet{Kulkarni2016HierarchicalDeepRL}, \citet{Andrychowicz2017HER}, and \citet{Florensa2018AutomaticGoalGeneration}. 

We trained a bottom-up agent with Proximal Policy Optimization (PPO) integrated with Monte Carlo Tree Search (MCTS) and Gumbel, and two top-down agents with PPO+MCTS and Soft Actor-Critic (SAC). We evaluate all three agents on two-variable polynomials over $\mathbb{F}_5$. The bottom-up agent achieves near-perfect success on small circuits (99.2\% aggregate up to complexity $C_3$), but performance degrades sharply, falling to $7.8\%$ at $C_{10}$ due to the combinatorial explosion of the action space. The top-down agents perform significantly better. PPO+MCTS reaches a $91.8\%$ success rate in finding optimal circuits across complexities $C_2$--$C_{10}$. On the other hand, the best SAC checkpoint achieved a success rate of 92.8\% at roughly one-fifth the compute cost. Both methods far exceed the $40\%$ uniform random baseline, demonstrating that the learned split and factor policies generalize to unseen targets.

We end the introduction by remarking that training RL agents to solve the arithmetic circuit problem with a high success rate is desirable for the following reasons. First, finding efficient circuits for important families of
polynomials, such as permanents, is of independent mathematical
interest and lies at the heart of algebraic complexity theory
\citep{Valiant1979Permanent,Burgisser1997AlgebraicComplexity,
ShpilkaYehudayoff2010ArithmeticCircuits}. Second, arithmetic circuit search provides a controlled testbed for learning symbolic search
heuristics, analogous to the role played by SAT solving as a benchmark
for combinatorial search and solver design
\citep{SATCompetition,Kurin2020GraphQSAT,
Yolcu2019LearningSAT}. Finally, the simplicity of the arithmetic
circuit environment makes it a useful model problem for more complicated
sequential circuit-construction tasks, including quantum circuit
synthesis and exact proof search in Lean
\citep{Moro2021QuantumCompilingRL,Fosel2021QuantumCircuitRL,
yang2023leandojo,DeepSeekProverV15}.

\section{The Arithmetic Circuit Problem}
\label{sec:arithmetic-circuit-problem}

In this section, we describe the two formulations of the arithmetic circuit problem that we considered. We introduce the concept of a \texttt{FactorLibrary} and discuss its implementation under both formulations.

\subsection{Background}
\label{subsec:background}

Let $\mathbb{F}_p$ be a finite field and let $X=\{x_0,\ldots,x_{n-1}\}$ be a set of variables. An arithmetic circuit over $\mathbb{F}_p[X]$ is a finite directed acyclic graph whose input gates are labeled by variables and constants and whose internal gates are labeled by addition $+$ or multiplication $\times$ \citep{Burgisser1997AlgebraicComplexity,ShpilkaYehudayoff2010ArithmeticCircuits}. The value at each internal gate is computed by performing the labeled operation on the values of the two incoming gates, and one gate is designated as the output gate. A circuit $C$ computes $f$ if the value at the output gate is equal to $f$. The size (or complexity) of a circuit $C$ is the total number of internal gates.

\begin{problem}[The arithmetic circuit problem]
Given a target polynomial $f\in\mathbb{F}_p[X]$, find a circuit $C$ with the minimum number of internal gates such that $C$ computes $f$.
\end{problem}

The arithmetic circuit problem is central to algebraic complexity theory. It is closely tied to Valiant's VP vs VNP conjecture, which is the algebraic analogue of P vs NP \citep{Valiant1979CompletenessClasses}. Finding efficient circuits for VNP-class polynomials such as permanents would be of great interest \citep{Valiant1979Permanent}. Even though this problem is hard in general, there are certain families of polynomials for which well-known efficient circuits exist. Horner's method \citep{Horner1819NewMethod} computes any degree-$d$ univariate polynomial using $2d$ gates, which is known to be optimal. Elementary symmetric polynomials and determinants satisfy similar recursions that produce efficient circuits. Moreover, a polynomial admits shorter-than-average circuits if it contains multiple factorizable subexpressions. Therefore, sums of products of polynomials are also a good heuristic for efficient circuits. For more, see \citet{Burgisser1997AlgebraicComplexity}. 

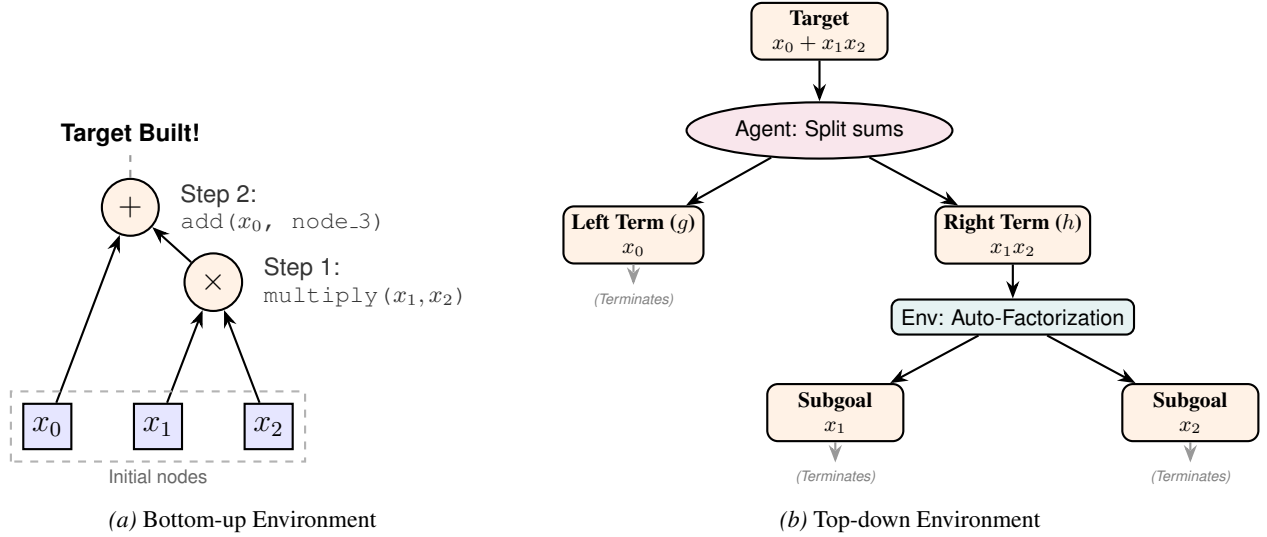
\begin{figure*}[t]
    \centering
    \begin{subfigure}[b]{0.43\textwidth}
        \centering
        \begin{tikzpicture}[
            node distance=1.1cm and 0.8cm,
            gate/.style={circle, draw=black, thick, fill=orange!10, minimum size=7mm, font=\large},
            leaf/.style={rectangle, draw=black, thick, fill=blue!10, minimum size=6mm, font=\large},
            edge/.style={->, >=Stealth, thick, draw=black},
            steptext/.style={text=black!80, font=\footnotesize\sffamily, align=left}
        ]
            \node[leaf] (x0) {$x_0$};
            \node[leaf] (x1) [right=of x0] {$x_1$};
            \node[leaf] (x2) [right=of x1] {$x_2$};
            
            \node[gate] (mul) [above=1.2cm of $(x1.north)!0.5!(x2.north)$] {$\times$};
            \draw[edge] (x1) -- (mul);
            \draw[edge] (x2) -- (mul);
            \node[right=0.15cm of mul, steptext] {Step 1:\\\texttt{multiply($x_1, x_2$)}};
            
            \node[gate] (add) [above=1.2cm of $(x0.north)!0.5!(mul.north)$] {$+$};
            \draw[edge] (x0) -- (add);
            \draw[edge] (mul) -- (add);
            \node[right=0.15cm of add, steptext] {Step 2:\\\texttt{add($x_0$, node\_3)}};
            
            \node[above=0.3cm of add, font=\small\sffamily\bfseries, color=black] (target) {Target Built!};
            \draw[dashed, thick, color=black!40] (add) -- (target);
            
            \draw[dashed, black!30, thick] ($(x0.south west)+(-0.15,-0.15)$) rectangle ($(x2.north east)+(0.15,0.15)$);
            \node[text=black!60, font=\scriptsize\sffamily] at ($(x1.south)+(0,-0.35)$) {Initial nodes};
        \end{tikzpicture}
        \caption{Bottom-up Environment}
        \label{fig:bottom-up}
    \end{subfigure}
    \hfill
    \begin{subfigure}[b]{0.54\textwidth}
        \centering
        \begin{tikzpicture}[
            scale=0.9,
            transform shape,
            node distance=0.7cm and 0.15cm,
            poly/.style={rectangle, draw=black, thick, fill=orange!10, rounded corners=4pt, minimum width=2.0cm, minimum height=0.7cm, align=center, font=\small},
            agent/.style={draw=black, thick, fill=purple!10, ellipse, align=center, inner sep=4pt, font=\small\sffamily},
            env/.style={draw=black, thick, fill=teal!10, rounded corners=3pt, rectangle, align=center, inner sep=4pt, font=\small\sffamily},
            edge/.style={->, >=Stealth, thick, draw=black},
            done/.style={text=black!50, font=\tiny\sffamily\itshape, align=center}
        ]
            \node[poly] (root) {\textbf{Target}\\$x_0 + x_1x_2$};
            
            \node[agent] (split) [below=0.6cm of root] {Agent: Split sums};
            \draw[edge] (root) -- (split);
            
            \node[poly] (g) [below left=0.8cm and 0.3cm of split] {\textbf{Left Term ($g$)}\\ $x_0$};
            \node[poly] (h) [below right=0.8cm and 0.3cm of split] {\textbf{Right Term ($h$)}\\ $x_1x_2$};
            \draw[edge] (split) -- (g);
            \draw[edge] (split) -- (h);
            
            \node[done] (g_done) [below=0.3cm of g] {(Terminates)};
            \draw[edge, dashed, draw=black!40] (g) -- (g_done);
            
            \node[env] (factor) [below=0.5cm of h] {Env: Auto-Factorization};
            \draw[edge] (h) -- (factor);
            
            \node[poly] (x1) [below left=0.7cm and -0.2cm of factor] {\textbf{Subgoal}\\$x_1$};
            \node[poly] (x2) [below right=0.7cm and -0.2cm of factor] {\textbf{Subgoal}\\$x_2$};
            \draw[edge] (factor) -- (x1);
            \draw[edge] (factor) -- (x2);
            
            \node[done] (x1_done) [below=0.3cm of x1] {(Terminates)};
            \node[done] (x2_done) [below=0.3cm of x2] {(Terminates)};
            \draw[edge, dashed, draw=black!40] (x1) -- (x1_done);
            \draw[edge, dashed, draw=black!40] (x2) -- (x2_done);
        \end{tikzpicture}
        \caption{Top-down Environment}
        \label{fig:top-down}
    \end{subfigure}
    
    \caption{Comparison of RL environment mechanics for finding arithmetic circuits for the target polynomial $f = x_0 + x_1x_2$.}
    \label{fig:framework_comparison}
\end{figure*}

\subsection{FactorLibrary and Recursive Subgoals}
\label{subsec:factor-library}

Polynomial arithmetic circuit optimization is closely related to algebraic expression optimization, program synthesis, library learning, and goal-directed reinforcement learning. Polynomial expressions can be optimized by combining algebraic factorization with common subexpression elimination \citep{Hosangadi2006OptimizingPolynomials}. More broadly, the idea of discovering reusable symbolic components has been studied in program synthesis and library learning. DreamCoder learns reusable abstractions across synthesis tasks through wake-sleep Bayesian program learning \citep{Ellis2020DreamCoder}, while LILO learns interpretable libraries by compressing and documenting code \citep{Grand2024LILO}. These works show that reusable symbolic components can improve search efficiency while preserving interpretability.

Our \texttt{FactorLibrary} follows a similar motivation, but adapts it to arithmetic circuit search. Instead of storing general program fragments, a \texttt{FactorLibrary} stores reusable polynomial factors or factorizable subexpressions that serve as subgoals across training episodes. Each library element is an explicit polynomial object, making the stored knowledge inspectable, verifiable, and directly meaningful in the target algebraic domain.

The \texttt{FactorLibrary} is also related to goal-directed and hierarchical reinforcement learning. Hierarchical Deep Reinforcement Learning uses intrinsic goals to decompose sparse-reward tasks \citep{Kulkarni2016HierarchicalDeepRL}. Hindsight Experience Replay improves sparse-reward learning by relabeling achieved states as goals \citep{Andrychowicz2017HER}. Automatic Goal Generation builds curricula of achievable but informative goals \citep{Florensa2018AutomaticGoalGeneration}. In contrast to these more general subgoal-generation methods, our \texttt{FactorLibrary} provides algebraically meaningful intermediate targets rather than learned latent goals or relabeled states.

It can be adapted to both the bottom-up and top-down formulations, although implementation details differ for each formulation. In both cases, the library acts as a form of symbolic memory: it records useful algebraic structures encountered or constructed during search and provides intermediate targets that guide the agent toward efficient circuit constructions. Compared to other subgoal techniques, the \texttt{FactorLibrary} has the advantage that each member is an honest polynomial, so the subgoals are fully explicit, interpretable, and verifiable.

\subsection{The Bottom-Up Formulation}
\label{subsec:bottom-up}

We adopt the formulation of the arithmetic circuit problem from \citet{Zhang2026CircuitBuilder}, which we refer to as the \textit{bottom-up} approach. Given a polynomial $f(x_0, \dots,x_{n-1})$, the initial game state consists of the variable gates $x_0, \dots, x_{n-1}$ and a constant gate $1$. Each subsequent game state at the $m^\text{th}$ iteration then consists of a partial circuit with gates $v_0,\ldots,v_{m-1}$. A valid action $\star\in\{+,\times\}$ connects two existing nodes $v_i,v_j$ and produces a new node $v_m=v_i\star v_j$. An episode succeeds when some node computes $f$. As the number of variables and the circuit size grows, the action space grows exponentially (see \citet[Section 3.2]{Zhang2026CircuitBuilder}), and non-optimized sequences of actions have very little chance of building a circuit which computes the target polynomial. Therefore, the bottom-up formulation is characterized by a sparse reward signal with large combinatorial growth. See \Cref{fig:framework_comparison} for a schematic of the bottom-up environment.

In the bottom-up formulation, we implement the \texttt{FactorLibrary} as an iteration-level short-term memory that stores reusable subexpressions. At the start of each episode, the target $f$ is factorized over $\mathbb{Z}$ using \texttt{SymPy}'s \texttt{factor\_list} and then reduced mod $p$. All nontrivial polynomial factors, those of degree at least one that are not base input nodes, are collected into the active subgoal set $\mathcal{F}_f$. During rollout, the first time a newly appended gate $u$ matches a member of $\mathcal{F}_f$, the agent receives a subgoal reward. If that subgoal is also present in the cross-episode library, an additional library bonus is awarded to encourage rediscovery of previously learned subexpressions.

We add two further kinds of reward signal to the library. Whenever $u$ is the newly added node, the residual $f - u$ is computed immediately and added to the library. In the case that $f$ is divisible by $u$, the quotient $f/u$ is added to the library. One-time completion bonuses are given when the circuit simultaneously contains $u$ and a node $v'$ satisfying $u + v' = f$ or $u \cdot v' = f$. Each bonus is given at most once per episode to ensure the primary success criterion remains exact target construction. After a successful episode, every non-input node of the circuit is entered into the library for future episodes. 

\subsection{The Top-Down Formulation}

\label{subsec:top-down}

We next introduce the \textit{top-down} formulation, which reverses the direction of search and regards the arithmetic circuit problem as a recursive decomposition problem. The initial state simply consists of the list $(f)$ where $f$ is the target polynomial. The valid actions are all two-part decompositions of the form $f = g + h$ for polynomials $g,h$. We refer to these as \textit{addition split points}. Once a split point is chosen, the game environment then immediately factors $g$ and $h$ and returns a list of factors $(g_1, \dots, g_r, \; h_1, \dots, h_s)$, and this list together with the chosen addition split point and multiplications in the factorizations becomes the next state. For each subsequent state $(f_1, \dots, f_l)$, the valid actions are addition split points for all factors, in the form $f_i = g_i + h_i$. The game terminates when the list consists of only single variables $x_j$ and constants $1$. See \Cref{fig:framework_comparison} for a schematic of the top-down environment. 

One advantage of this formulation lies in its deterministic nature. The agent is not asked to guess the next gate in a partial circuit. Instead, the agent chooses how to decompose the remaining target. Since factorization algorithms over finite fields are exact, the agent is always \textit{guaranteed} to build a valid circuit for $f$ given enough action steps. 

In the top-down formulation, the \texttt{FactorLibrary} stores factorizable polynomials, specifically those for which building the polynomial via its factorization takes strictly fewer gates than direct construction. The library is prebuilt with examples such as $(x_i + a)^2$ for each variable $x_i$ and each nonzero constant $a \in \mathbb{F}_p$, $(x_i + a)(x_j + b)$ for each variable pair and all constants $a, b \in \mathbb{F}_p$, and elementary symmetric polynomials up to degree three. As search proceeds, every factorizable polynomial is included, so the library grows dynamically with the agent's sequence of actions. 

Before the agent chooses a split point on a polynomial $f$ in the current state, the library identifies all entries $g$ whose coefficients are strictly smaller than those of $f$, so that $g$ can serve as an additive component. The agent gains a reward, up to a scaling factor, if it chooses an action that hits a library entry. See \Cref{fig:fl-top-down} for an illustration. 


Another advantage of the top-down environment is that it has a much smaller effective action space for fixed circuit complexity. If a size-$C$ circuit has $A$ addition gates and $M$ multiplication gates, with $C=A+M$, then the bottom-up agent must search over all $C$ gate choices. In contrast, the top-down agent only chooses additive split points, whereas multiplication gates are supplied by deterministic factorization and not part of the action sequence. The number of top-down action sequences is at most $K^A$ where $K$ is the total number of candidate splits, rather than the space of all circuits of complexity $C$. Since typically $A\leq C$ and often $A\ll C$ for factorizable targets, this gives a substantial reduction in effective search space. Indeed, this dramatic reduction in the size of the action space has led to superior training trajectories for the top-down agents compared to the bottom-up ones, as we discuss below. 

\section{Training}
\label{sec:training}

We now describe the training pipelines for the two formulations introduced in \Cref{sec:arithmetic-circuit-problem}. For bottom-up, we adopted \texttt{CircuitBuilder}'s gate-action environment and added JAX-\texttt{mctx} Gumbel search, PPO updates, and \texttt{FactorLibrary} shaping. We only considered PPO, because \citet{Zhang2026CircuitBuilder} points out that PPO performs better at higher complexities. For top-down, we train a policy over variable-sized candidate decompositions, using factorization, memoization, \texttt{FactorLibrary} and AND/OR MCTS to evaluate recursive subgoals. We compared the efficacy of PPO and SAC in this new environment. All built polynomials are checked with exact arithmetic over $\mathbb F_p$. We summarize both training pipelines in \Cref{fig:training-pipelines}.

\subsection{Bottom-Up Gumbel-PPO-MCTS}

In the bottom-up environment, each partial circuit is represented as a DAG whose nodes store finite-field polynomials. Observations contain a padded circuit graph, node features, the target coefficient vector, and a valid-action mask. Actions are encoded by an operation $\star\in\{+,\times\}$ together with an upper-triangular pair of existing nodes $(v_i,v_j)$, producing the new node $v_i\star v_j$. A graph encoder embeds the partial circuit, a target encoder embeds the coefficient vector of the target polynomial, and the resulting representation is passed to policy and value heads.

The implementation is built on JAX, with the network written in Flax and optimization handled by optax with gradient-clipped Adam. The environment is also implemented in pure JAX as a collection of fixed-size arrays (node coefficient matrix, edge arrays, subgoal arrays, and target vector), enabling JIT compilation of the full step function. All $B$ parallel environments are stepped in a single JIT-compiled call, and the \texttt{FactorLibrary} reward path, including subgoal coefficient comparisons, library-known flag lookups, and completion checks, is executed as JAX operations over preloaded arrays, keeping the entire training loop GPU-resident. Batched MCTS uses Google DeepMind's \texttt{mctx} library, which searches all $B$ environments simultaneously on GPU.

At each rollout step, the policy is locally improved by root-only Gumbel search \citep{Danihelka2022GumbelPlanning}. The network supplies masked logits and a value estimate. Gumbel-Top-$k$ sampling selects a bounded set of promising root actions, and Sequential Halving compares those candidates by shallow lookahead. The search returns both a selected gate action and a completed-$Q$ root distribution. PPO then updates the network using the clipped surrogate objective \citep{schulman2017ppo}, with advantages computed by generalized advantage estimation \citep{Schulman2016GAE}. The stored network log-probability, not the MCTS probability, is used in the PPO ratio. The Gumbel-improved root distribution is used as an auxiliary cross-entropy distillation target, following the expert-iteration pattern in which search improves local decisions and the network generalizes these improved decisions across all states \citep{Anthony2017ExpertIteration,Silver2018AlphaZero}.

The \texttt{FactorLibrary} operates on the level of reward shaping. At episode reset, nontrivial factors of the remaining target are added as episode subgoals. During rollout, we give a subgoal reward to a newly constructed gate if it is a first-time match for a library subgoal, and a library bonus if that subgoal was previously constructed in a successful episode. If the new node $u$ and an existing node $v$ satisfy $u+v=f$ or $uv=f$, the transition receives a completion bonus. These bonuses are bounded and one-time per episode, so the success criterion remains tied to exactly constructing the target. The \texttt{FactorLibrary}, however, persists across all episodes and serves as the short-term memory of the model. See \Cref{alg:bottom-up-gumbel} for pseudocode, \Cref{fig:bottom-up} and \Cref{fig:fl-bottom-up} for illustrations. 

\begin{algorithm}[t]
\caption{Bottom-Up Gumbel-PPO-MCTS with \texttt{FactorLibrary}}
\label{alg:bottom-up-gumbel}
\footnotesize
\begin{algorithmic}
\REQUIRE target distribution $\mathcal D_{\mathrm{tar}}$, budget $B$, network $(\pi_\theta,V_\theta)$, library $\mathcal L$, search budget $K$
\FOR{each training episode}
    \STATE Sample target $f\sim\mathcal D_{\mathrm{tar}}$ and reset circuit $C\gets\{1,x_0,\ldots,x_{n-1}\}$
    \STATE Initialize unrewarded subgoals $\mathcal F_f\gets \mathrm{NontrivialFactors}(f)$
    \FOR{$t=0,\ldots,B-1$}
        \STATE Encode $s_t=(C,f,\mathcal F_f,\mathcal L)$ with valid gate-action mask
        \STATE $(a_t,\widehat{\pi}_t)\gets \textsc{GumbelRootSearch}(s_t,\pi_\theta,V_\theta,K)$
        \STATE Append $u\gets v_i\star v_j$ to $C$ for $a_t=(\star,i,j)$
        \STATE Set $r_t$ by exact terminal, subgoal, library, and one-step completion checks
        \STATE Remove any newly rewarded subgoals from $\mathcal F_f$
        \STATE Store $(s_t,a_t,r_t,\widehat{\pi}_t,V_\theta(s_t),\log\pi_\theta(a_t\mid s_t))$
        \IF{$u=f$}
            \STATE Add constructed non-input nodes of $C$ to $\mathcal L$ and terminate episode
        \ENDIF
    \ENDFOR
    \STATE Update $\theta$ by PPO with GAE and distillation to $\widehat{\pi}_t$
\ENDFOR
\end{algorithmic}
\end{algorithm}

\subsection{Top-Down PPO+MCTS and SAC}

The top-down implementation trains on the decomposition environment described in \Cref{subsec:top-down}. At each decision point, the active polynomial is the first element in the list of unresolved subgoals. The environment generates a variable-sized candidate set consisting of additive splits and, when the active polynomial factors nontrivially, a whole-polynomial factor action. Candidate splits are generated from term supports, random masks, common-factor clusters, known patterns, and \texttt{FactorLibrary} matches. The policy scores the current candidate list generated by the environment.

After an action is selected, the environment factors the split pieces over $\mathbb F_p$. Only additive splits are counted towards agent actions and penalized in reward shaping. Base cases and memoized subproblems are resolved immediately, and unresolved nontrivial factors are pushed back onto the game state. The step reward is the reduction of cost before and after the action compared to the baseline cost given by random actions and heuristic search, with an optional library-match bonus. 

Planning is performed by memoized AND/OR PUCT search, building on UCT and MCTS \citep{Kocsis2006UCT,Browne2012MCTS}. OR nodes are candidate actions for a polynomial. AND expansions are deterministic: after an action is chosen, factorization determines all child subproblems that must be solved. Search backs up total circuit cost rather than the value on a single leaf node. Our use of priors and values follows the AlphaZero search regime \citep{Silver2018AlphaZero}.

In PPO+MCTS mode, MCTS runs at each action step. The distribution of root visits is used for action selection and as a cross-entropy distillation target in the PPO loss. We follow \citet{schulman2017ppo} for clipped surrogate, value loss, and entropy bonus. In SAC mode, the critic scores each candidate decomposition action by its expected entropy-regularized return, and the actor is trained with a discrete-action maximum-entropy objective \citep{Haarnoja2018SAC,Christodoulou2019DiscreteSAC}. See \Cref{alg:top-down-clean} for pseudocode.

\begin{algorithm}[t]
\caption{Top-Down PPO+MCTS and SAC with \texttt{FactorLibrary}}
\label{alg:top-down-clean}
\footnotesize
\begin{algorithmic}
\REQUIRE target distribution $\mathcal D_{\mathrm{tar}}$, budget $H$, model $(\pi_\theta,V_\theta)$, optional critic $Q_\phi$, library $\mathcal L$, search budget $K$
\FOR{each training episode}
    \STATE Sample target $f\sim\mathcal D_{\mathrm{tar}}$
    \STATE Reset list of subgoals $\mathcal Q\gets(f)$, trace $\mathcal T\gets\emptyset$, memo table $M\gets\emptyset$
    \FOR{$t=0,\ldots,H-1$}
        \IF{$\mathcal Q=\emptyset$}
            \STATE Reconstruct and verify circuit from $\mathcal T$; add useful subtraces to $\mathcal L$
            \STATE Terminate episode
        \ENDIF
        \STATE Let $q$ be the active unresolved polynomial
        \STATE Generate verified actions $\mathcal A(q)$: splits, library matches, and possible factor action
        \STATE $(a_t,\widehat{\pi}_t)\gets \textsc{AndOrMCTS}(q,\mathcal A(q),M,\pi_\theta,V_\theta,K)$
        \STATE Execute $a_t$: factor selected pieces, charge rebuild cost, and update $\mathcal Q,\mathcal T,M$
        \STATE Set reward to baseline cost saving plus any verified library reward
        \STATE Store transition and search target $\widehat{\pi}_t$ in rollout or replay buffer
    \ENDFOR
    \STATE Update by PPO with MCTS distillation, or by SAC from replay using candidate-set $Q$-values
\ENDFOR
\end{algorithmic}
\end{algorithm}

\section{Results and Comparisons}
\label{sec:results}

In this section, we discuss our training and evaluation results. For the bottom-up branch, we analyze the success rate of building a circuit for the target polynomial across the training curriculum, which is based on increasing heuristic size of the circuits. For the top-down branch, we analyze the success rate of the agent beating the baseline and achieving the optimal circuit length across circuit sizes. 

\subsection{Bottom-Up Gumbel-PPO-MCTS}

The bottom-up Gumbel-PPO-MCTS run uses a JAX implementation with \texttt{mctx} Gumbel MuZero search. \texttt{FactorLibrary} provides reward shaping. We trained using a curriculum based on circuit size, over two-variable targets in \(\mathbb F_5\). Let $C_i$ denote circuits of heuristic size $i$, meaning circuits generated from $i$ random gates. The curriculum proceeds through adjacent complexity buckets, starting from \(C_1\), then \(C_1,C_2\), then \(C_2,C_3\), and continuing up to \(C_9,C_{10}\). We trained the agent using a \(4\)-layer GNN with hidden dimension \(384\), batch size \(1024\), \texttt{mctx} Gumbel MuZero search with \(64\) simulations and \(16\) actions, on a single NVIDIA RTX PRO 6000 Blackwell GPU. A condensed summary of the curriculum is given in \Cref{tab:bottomup-condensed}; the full phase-by-phase results are in \Cref{tab:bottomup-gumbel-curriculum} in the appendix.

The results show that Gumbel search and the \texttt{FactorLibrary} make the learning effective at low circuit size. The agent had \(99.2\%\) aggregate success up to \(C_3\), showing that the network has memorized the constructions of most small-size circuits. Performance remains strong through \(C_4\), but begins to degrade at \(C_5\), and the drop becomes sharper at higher complexities. The \(C_5,C_6\) phase ends at \(45.0\%\), the \(C_6,C_7\) phase at \(26.8\%\), and the \(C_9,C_{10}\) phase at \(7.8\%\). This is the expected failure mode of the primitive gate-action formulation, which coincides with the combinatorial explosion of the search space. Even with improved root search and factor-library rewards, the agent still struggles to discover useful intermediate expressions inside a rapidly expanding action space. 

\begin{table}[h]
\centering
\caption{Bottom-up Gumbel-PPO-MCTS curriculum (condensed). Seven representative phases illustrate the learning trajectory. Full results in \Cref{tab:bottomup-gumbel-curriculum}.}
\label{tab:bottomup-condensed}
\small
\setlength{\tabcolsep}{5pt}
\begin{tabular}{@{}clc@{}}
\toprule
Phase & Train buckets & Agg.\ success \\
\midrule
1  & $C_1$          & 93.7\% \\
3  & $C_2,C_3$      & 99.2\% \\
4  & $C_3,C_4$      & 91.7\% \\
5  & $C_4,C_5$      & 71.7\% \\
6  & $C_5,C_6$      & 45.0\% \\
7  & $C_6,C_7$      & 26.8\% \\
10 & $C_9,C_{10}$   & 7.8\%  \\
\bottomrule
\end{tabular}
\end{table}

\subsection{Top-Down PPO+MCTS versus SAC}

We trained the top-down PPO+MCTS agent with an MLP with $3$ hidden layers of width $128$ on an ensemble of targets combining Horner schemes, sums of products, and elementary symmetric polynomials. The training ran for \(1000\) iterations and used \(16\) rollouts per update, \(16\) candidate actions per step, maximum episode length \(24\), and \(48\) MCTS simulations of depth \(6\) on an RTX 4070 Ti.  The top-down SAC agent used the same actor architecture as PPO but replaced MCTS with off-policy discrete SAC using twin \(3\)-layer width-\(128\) critics, replay batch size \(64\), \(16\) gradient steps per iteration, Polyak target updates, and an automatically tuned entropy temperature with target entropy scale \(0.4\).

For evaluation, we used a fixed \(450\)-target training set for in-distribution tracking and a disjoint \(207\)-target held-out set for the reported generalization results, across \(C_2,\ldots,C_{10}\). For \(C_2,\ldots,C_8\), the \(C_k\) labels are computed by exact search over the split/factor/direct move set. The \(C_9\) and \(C_{10}\) buckets are partly heuristic-labeled, so verified-optimality claims are limited to \(C_2,\ldots,C_8\). See \Cref{tab:topdown-summary} for success rates and \Cref{fig:diag-heldout-png} for training curves.

\begin{table}[h]
\centering
\caption{Top-down evaluation metrics. }
\label{tab:topdown-summary}
\small
\setlength{\tabcolsep}{4pt}
\begin{tabular}{@{}lccccc@{}}
\toprule
Method & Final & Best & Train & Gap & Time \\
\midrule
PPO+MCTS & 0.918 & 0.918 & 0.918 & 0.01 & 2.5 h \\
SAC      & 0.889 & 0.928 & 0.844 & 0.05 & 30 m \\
Random   & 0.40  & --    & --    & --   & -- \\
\bottomrule
\end{tabular}
\end{table}

Both top-down methods substantially outperform the random baseline. PPO+MCTS reaches a final held-out \(C_k\)-match rate of \(0.918\) ($k \le 8$), compared with \(0.40\) for uniform random action selection over the same candidate sets. This indicates that the learned split/factor policy generalizes to unseen targets rather than memorizing the training set. SAC is competitive despite not using MCTS. Its final held-out success rate is \(0.889\), and its best checkpoint reaches \(0.928\), slightly above the final PPO+MCTS checkpoint. The SAC method is much cheaper computationally, as each run takes about 30 minutes on an RTX 4070 Ti, compared with about 2.5 hours for PPO+MCTS. However, it appears less stable due to drift late in training. The choice between the two methods is therefore stability versus compute.

\subsection{Limitations}

The scale of our experiments is restricted to few variables, coefficients in $\mathbb F_5$, and relatively small circuit complexity. In the bottom-up environment, the action space grows quadratically in the number of variables $n$ and complexity $C$, and the total search space grows super-exponentially in $n$ and $C$ \citep{Zhang2026CircuitBuilder}.  The top-down action space can be understood in terms of the number of split points. An active polynomial of degree at most $d$ in $n$ variables over $\mathbb F_p$ admits $p^{\binom{n+d}{n}}$ ordered additive decompositions $f=g+h$, since $g$ ranges freely over that space and the degree satisfies $d\le 2^C$. The growth of the search space in both environments makes training lengthy, and our experiments are constrained by available computing resources. Additional work is necessary to extend our methods to highly complex polynomials.

\section{Conclusion and Future Directions}

We studied two reinforcement learning formulations for the arithmetic circuit problem and implemented \texttt{FactorLibrary} in each. The bottom-up formulation extends gate-level search with Gumbel-PPO-MCTS and \texttt{FactorLibrary} reward shaping, implemented in JAX with \texttt{mctx} for GPU-batched MCTS. The top-down formulation searches over recursive additive split points, using memoized AND/OR MCTS and a JAX-accelerated \texttt{FactorLibrary} to produce candidate actions. 

Our results show that top-down agents perform substantially better at higher circuit complexities. The challenge for bottom-up is that MCTS search performance degrades sharply beyond $C_5$, consistent with the combinatorial explosion of the action space. The top-down agents, which are fed algebraic subgoals directly rather than individual gates, maintain strong performance through $C_{10}$. The SAC variant achieves competitive results at a fraction of the compute, suggesting that the top-down environment is well suited for off-policy learning. The \texttt{FactorLibrary} can be well-adapted to both formulations. It stores reusable subgoals that are certified and fully interpretable, enabling sophisticated reward shaping across multiple episodes.

In future work, we plan to scale to more variables and higher degree targets, more carefully separate training-time from inference-time compute, and identify which polynomial families favor each formulation. We hope to use our trained models to discover efficient circuits for permanents and other VNP-hard families of polynomials. In a follow-up work, we will study a two-player formulation of the top-down environment. The \texttt{FactorLibrary} mechanism also alludes to a domain-specific analogue of a ``lemma library'' in the context of auto-proof generation. The arithmetic circuit problem provides a controlled setting for studying how learned systems acquire, reuse, and generalize symbolic subgoals.

\section*{Acknowledgements}
This project is a part of the UW Math AI Lab. We thank the UW eScience School for computing resources. CPU and GPU computing were in part done using AWS credits from the UW eScience School and UW IT, and also in part done using the UW Research Computing Club funded from the UW Student Technology Fee Committee. Parts of the code base and diagrams in the paper were produced with the help of ChatGPT 5.5 and Claude Opus 4.7.

\bibliography{references}
\bibliographystyle{icml2026}

\appendix

\onecolumn

\newpage

\section{Results and Plots from Experiments}
\label{app:additional-results}

\begin{figure}[htbp]
    \centering
    \includegraphics[width=\linewidth]{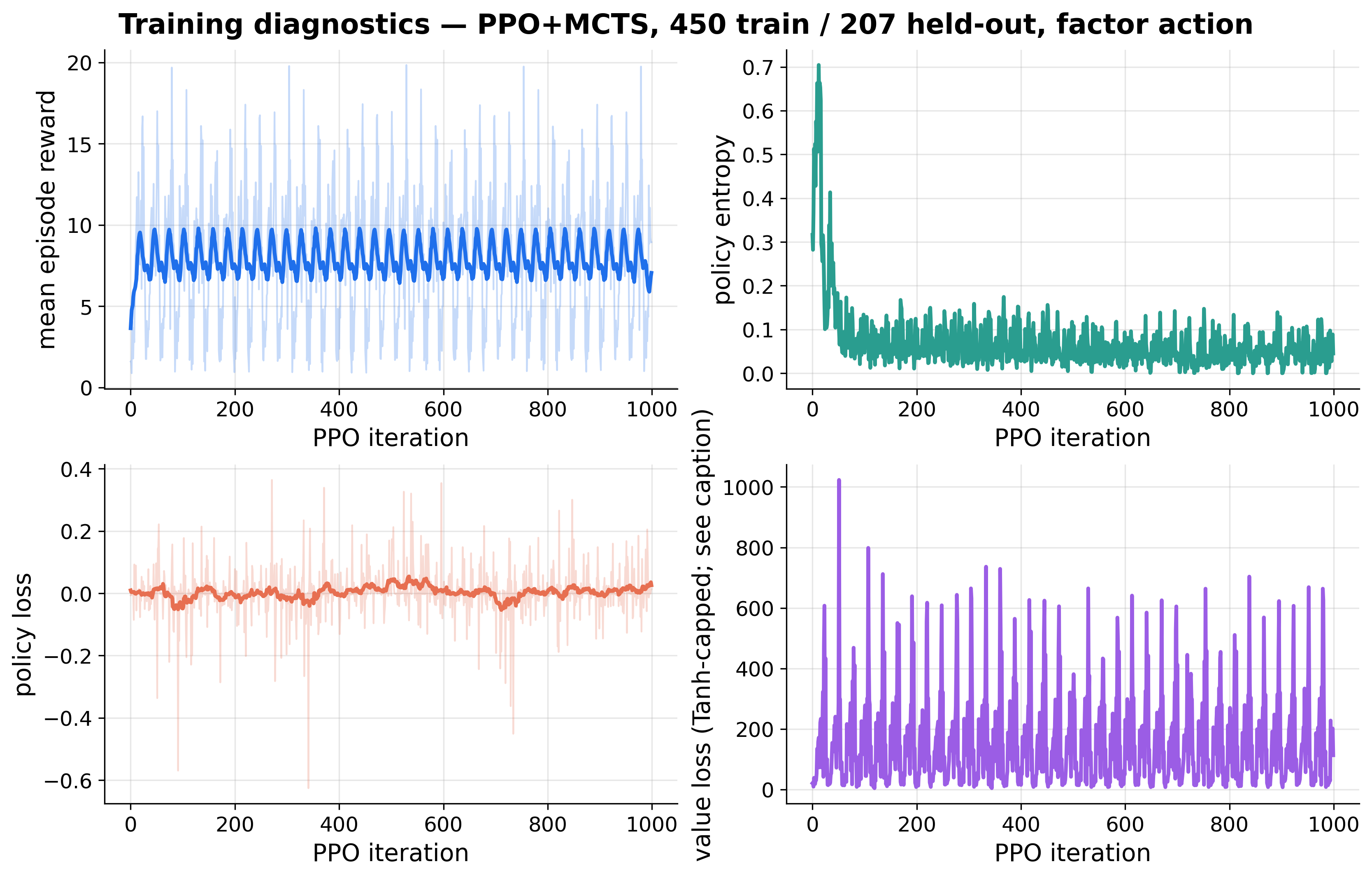}
    \caption{Training curves.}
    \label{fig:diag-training-curves}
\end{figure}

\begin{figure}[htbp]
    \centering
    \includegraphics[width=\linewidth]{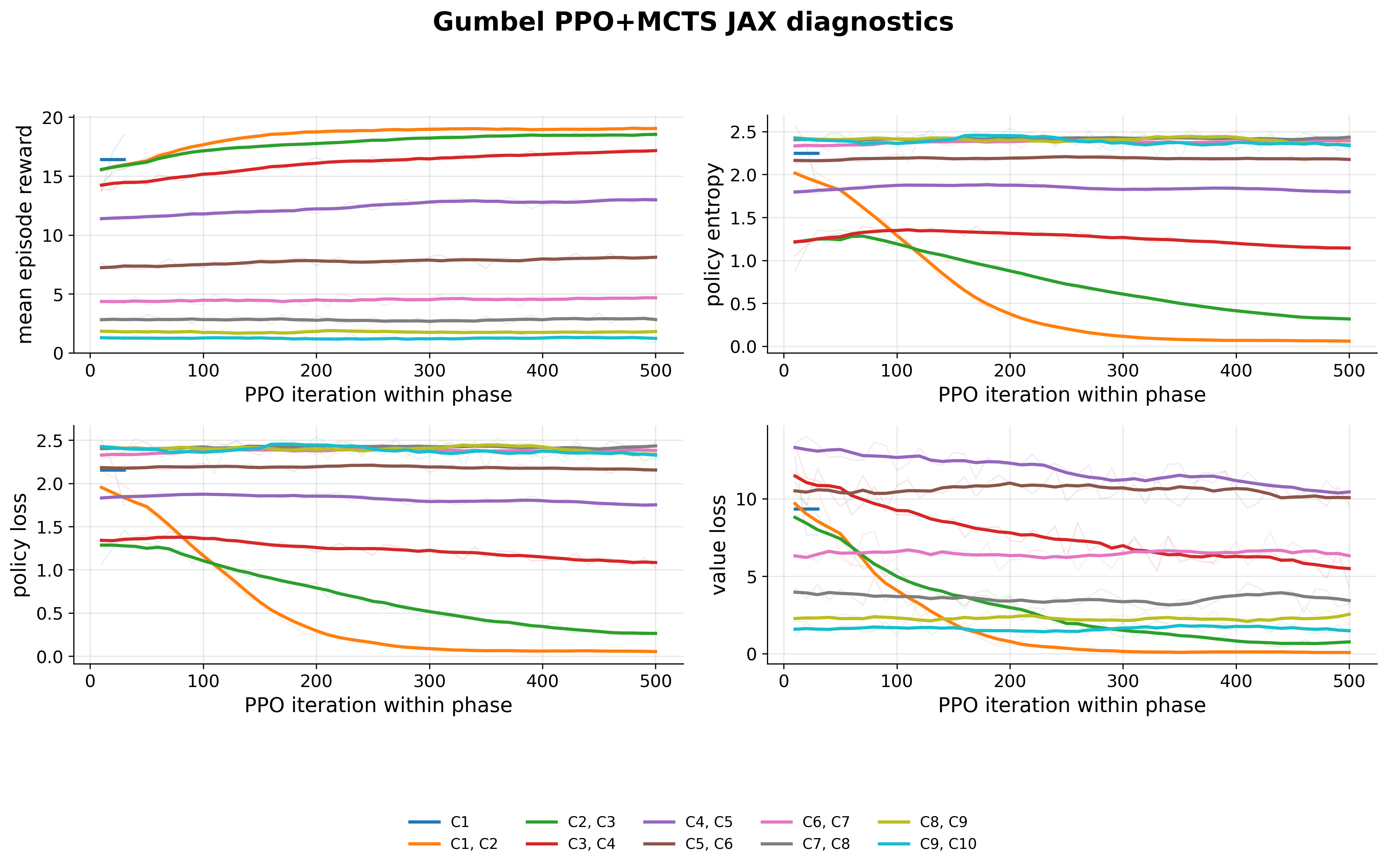}
    \caption{Overlaid diagnostics for the Gumbel curriculum phases.}
    \label{fig:diag-gumbel-diagnostics}
\end{figure}

\begin{figure}[htbp]
    \centering
    \includegraphics[width=\linewidth]{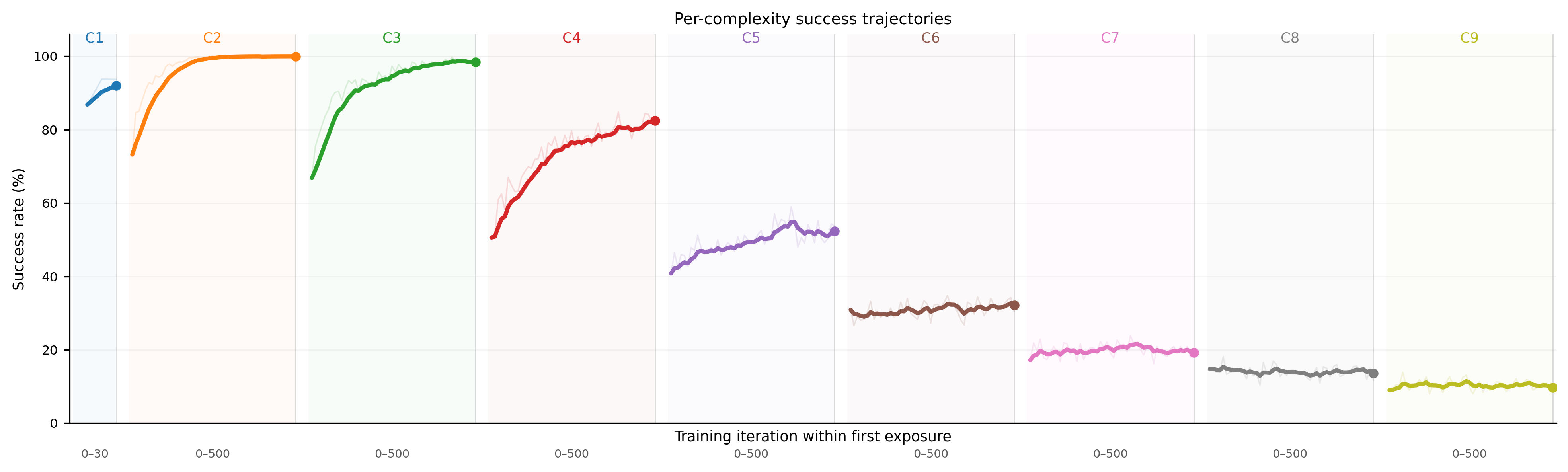}
    \caption{Staggered success rates across curriculum phases.}
    \label{fig:diag-stagger-sr}
\end{figure}

\begin{figure}[htbp]
    \centering
    \includegraphics[width=\linewidth]{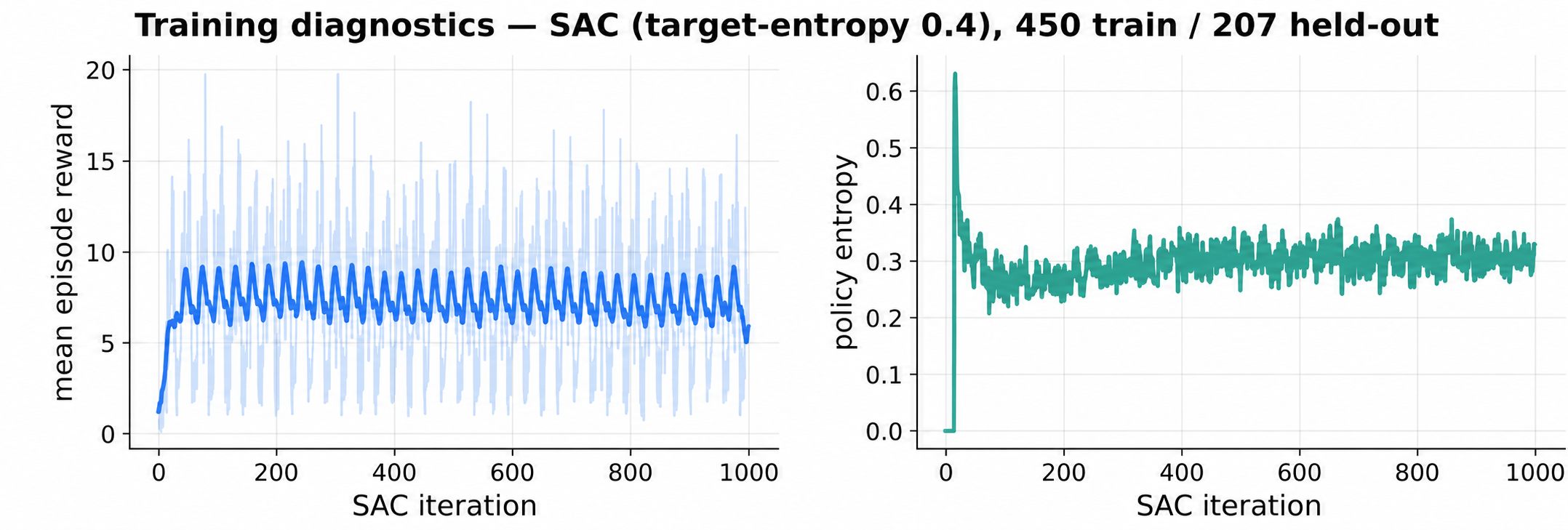}
    \caption{SAC training diagnostics for the top-down run.}
    \label{fig:diag-screenshot}
\end{figure}

\begin{figure}[htbp]
    \centering
    \includegraphics[width=\linewidth]{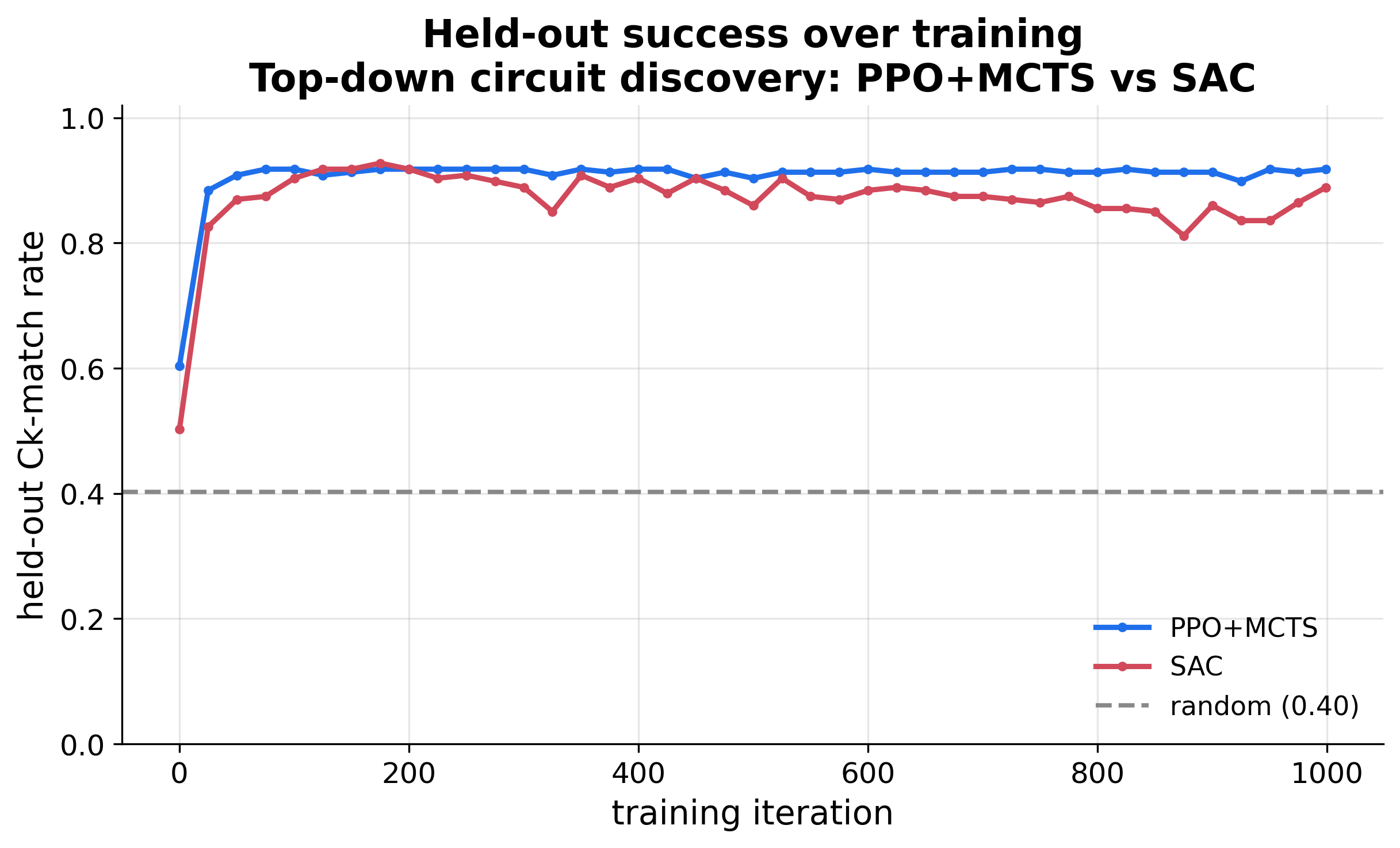}
    \caption{Held-out \(C_k\)-match rate over training iterations.}
    \label{fig:diag-heldout-png}
\end{figure}

\Cref{tab:bottomup-gumbel-curriculum} gives the complete phase-by-phase curriculum for the bottom-up Gumbel-PPO-MCTS run. Each phase is initialized from the previous checkpoint. Library size tracks the number of distinct polynomials accumulated in the \texttt{FactorLibrary} across episodes; the peak at Phase~7 ($15{,}051$ entries) reflects the transition from regimes where many circuits are solved (and thus registered) to regimes where the agent rarely succeeds. Figures~\ref{fig:diag-training-curves}--\ref{fig:diag-heldout-png} show training curves and final per-complexity breakdowns for the top-down agents.

\begin{table*}[htbp]
    \centering
    \caption{Bottom-up JAX Gumbel-PPO-MCTS full curriculum. Each phase trains on the indicated complexity bucket(s) and is initialized from the previous phase. Success is exact target-construction rate in the gate-action environment.}
    \label{tab:bottomup-gumbel-curriculum}
    \small
    \setlength{\tabcolsep}{6pt}
    \begin{tabular}{@{} cl c c l r r @{}}
        \toprule
        \textbf{Phase} & \textbf{Train Buckets} & \textbf{Iter.} & \textbf{Agg. Success} & \textbf{Per-Bucket Success} & \textbf{Lib. Size} & \textbf{Reward} \\
        \midrule
        1  & $C_1$        & 30  & 93.65\% & $C_1=93.7\%$               & 501   & 18.535 \\
        2  & $C_1,C_2$    & 500 & 99.90\% & $C_1=100.0\%,\ C_2=99.8\%$ & 959   & 19.263 \\
        3  & $C_2,C_3$    & 500 & 99.22\% & $C_2=100.0\%,\ C_3=98.4\%$ & 1368  & 18.696 \\
        4  & $C_3,C_4$    & 500 & 91.70\% & $C_3=100.0\%,\ C_4=83.4\%$ & 3985  & 17.240 \\
        5  & $C_4,C_5$    & 500 & 71.68\% & $C_4=89.6\%,\ C_5=53.7\%$  & 8839  & 13.137 \\
        6  & $C_5,C_6$    & 500 & 45.02\% & $C_5=59.6\%,\ C_6=30.5\%$  & 13693 & 7.973  \\
        7  & $C_6,C_7$    & 500 & 26.76\% & $C_6=35.2\%,\ C_7=18.4\%$  & 15051 & 4.686  \\
        8  & $C_7,C_8$    & 500 & 17.19\% & $C_7=22.7\%,\ C_8=11.7\%$  & 11625 & 2.786  \\
        9  & $C_8,C_9$    & 500 & 12.40\% & $C_8=16.0\%,\ C_9=8.8\%$   & 8860  & 1.830  \\
        10 & $C_9,C_{10}$ & 500 & 7.81\%  & $C_9=8.4\%,\ C_{10}=7.2\%$ & 7279  & 0.925  \\
        \bottomrule
    \end{tabular}
\end{table*}

\begin{figure*}[p]
\centering
\begin{subfigure}[t]{0.48\textwidth}
\centering
\begin{tikzpicture}[
    node distance=0.62cm,
    box/.style={rectangle, draw=black, thick, rounded corners=3pt,
        align=center, minimum width=3.0cm, minimum height=0.62cm,
        fill=blue!6, font=\scriptsize},
    search/.style={rectangle, draw=black, thick, rounded corners=3pt,
        align=center, minimum width=3.0cm, minimum height=0.62cm,
        fill=orange!12, font=\scriptsize},
    data/.style={cylinder, shape border rotate=90, draw=black, thick,
        align=center, minimum width=2.4cm, minimum height=0.62cm,
        fill=purple!8, font=\scriptsize},
    arr/.style={->, thick, >=Stealth}
]
\node[box] (init) {Partial circuit\\$\{1,x_0,\ldots,x_{n-1}\}$};
\node[data, below=of init] (lib) {\texttt{FactorLibrary}};
\node[box, below=of lib] (obs) {Graph, target vector, mask};
\node[box, below=of obs] (net) {Policy-value network};
\node[search, below=of net] (search) {Gumbel root search};
\node[box, below=of search] (gate) {Append gate $v_i\star v_j$};
\node[box, below=of gate] (checks) {Success and subgoal checks};
\node[box, below=of checks] (update) {PPO update + distillation};

\draw[arr] (init) -- (lib);
\draw[arr] (lib) -- (obs);
\draw[arr] (obs) -- (net);
\draw[arr] (net) -- (search);
\draw[arr] (search) -- (gate);
\draw[arr] (gate) -- (checks);
\draw[arr] (checks) -- (update);
\draw[arr] (update.east) -- ++(0.55,0) |- (net.east);
\draw[arr] (checks.west) -- ++(-0.45,0) |- (lib.west);
\end{tikzpicture}
\caption{Bottom-up pipeline.}
\label{fig:bottom-up-training-pipeline}
\end{subfigure}
\hfill
\begin{subfigure}[t]{0.48\textwidth}
\centering
\begin{tikzpicture}[
    node distance=0.62cm,
    box/.style={rectangle, draw=black, thick, rounded corners=3pt,
        align=center, minimum width=3.0cm, minimum height=0.62cm,
        fill=blue!6, font=\scriptsize},
    search/.style={rectangle, draw=black, thick, rounded corners=3pt,
        align=center, minimum width=3.0cm, minimum height=0.62cm,
        fill=orange!12, font=\scriptsize},
    data/.style={cylinder, shape border rotate=90, draw=black, thick,
        align=center, minimum width=2.4cm, minimum height=0.62cm,
        fill=purple!8, font=\scriptsize},
    arr/.style={->, thick, >=Stealth}
]
\node[box] (frontier) {List of subgoals $\mathcal Q=(f)$};
\node[data, below=of frontier] (lib) {\texttt{FactorLibrary}};
\node[box, below=of lib] (cand) {Candidate actions\\splits + factor action};
\node[box, below=of cand] (score) {Candidate scoring};
\node[search, below=of score] (mcts) {Memoized AND/OR MCTS};
\node[box, below=of mcts] (factor) {Execute action and factor pieces};
\node[box, below=of factor] (updatefrontier) {Update list of subgoals, trace, memo};
\node[box, below=of updatefrontier] (learn) {PPO/SAC update + search targets};

\draw[arr] (frontier) -- (lib);
\draw[arr] (lib) -- (cand);
\draw[arr] (cand) -- (score);
\draw[arr] (score) -- (mcts);
\draw[arr] (mcts) -- (factor);
\draw[arr] (factor) -- (updatefrontier);
\draw[arr] (updatefrontier) -- (learn);
\draw[arr] (learn.east) -- ++(0.95,0) |- (score.east);
\draw[arr] (updatefrontier.west) -- ++(-0.45,0) |- (lib.west);
\end{tikzpicture}
\caption{Top-down pipeline.}
\label{fig:top-down-training-pipeline}
\end{subfigure}
\caption{Training pipelines for bottom-up and top-down environments.}
\label{fig:training-pipelines}
\end{figure*}
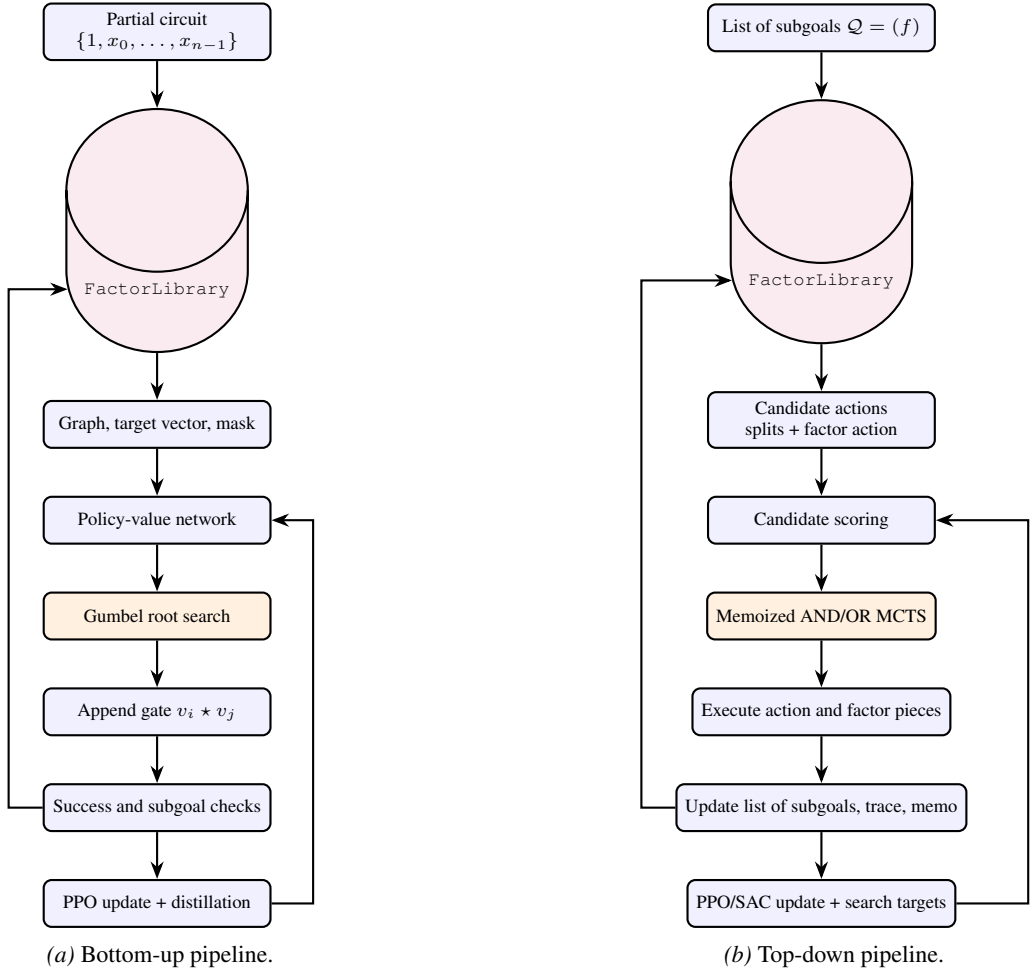

\begin{figure*}[t]
\centering
\begin{subfigure}[t]{0.46\textwidth}
\centering
\begin{tikzpicture}[
    node distance=0.62cm,
    box/.style={rectangle, draw=black, thick, rounded corners=3pt,
        align=center, minimum width=3.0cm, minimum height=0.62cm,
        fill=blue!6, font=\scriptsize},
    rew/.style={rectangle, draw=black, thick, rounded corners=3pt,
        align=center, minimum width=3.0cm, minimum height=0.62cm,
        fill=green!8, font=\scriptsize},
    data/.style={cylinder, shape border rotate=90, draw=black, thick,
        align=center, minimum width=2.4cm, minimum height=0.62cm,
        fill=purple!8, font=\scriptsize},
    arr/.style={->, thick, >=Stealth},
    sidelab/.style={text=black!55, font=\tiny\sffamily\itshape, align=center},
]
\node[box] (sample)  {Sample target $f$};
\node[box,  below=of sample]  (factor) {Factorize $f$ over $\mathbb{Z}$ (SymPy)};
\node[data, below=of factor]  (sg)     {Episode subgoals $\mathcal{F}_f$};
\node[box,  below=of sg]      (gate)   {New gate $u = v_i \star v_j$};
\node[rew,  below=of gate]    (sgrew)  {$u \in \mathcal{F}_f$: subgoal reward};
\node[rew,  below=of sgrew]   (librew) {$u \in \mathcal{L}$: library bonus};
\node[rew,  below=of librew]  (comp)   {$u+v'=f$ or $u \cdot v'=f$:\\completion bonus};
\node[data, below=of comp]    (lib)    {\texttt{FactorLibrary} $\mathcal{L}$};
\draw[arr] (sample) -- (factor);
\draw[arr] (factor) -- (sg);
\draw[arr] (sg)     -- (gate);
\draw[arr] (gate)   -- (sgrew);
\draw[arr] (sgrew)  -- (librew);
\draw[arr] (librew) -- (comp);
\draw[arr] (comp.east) -- ++(0.55,0)
    node[sidelab, right, text width=0.9cm] {on\\success}
    |- (lib.east);
\draw[arr, dashed, draw=black!45] (lib.west) -- ++(-0.55,0) |- (librew.west);
\draw[arr, dashed, draw=black!45] (lib.west) -- ++(-0.55,0) |- (sg.west)
    node[sidelab, left, pos=0.5, text width=1.0cm] {next\\episode};
\end{tikzpicture}
\caption{Bottom-up \texttt{FactorLibrary}.}
\label{fig:fl-bottom-up}
\end{subfigure}
\hfill
\begin{subfigure}[t]{0.46\textwidth}
\centering
\begin{tikzpicture}[
    node distance=0.62cm,
    box/.style={rectangle, draw=black, thick, rounded corners=3pt,
        align=center, minimum width=3.0cm, minimum height=0.62cm,
        fill=blue!6, font=\scriptsize},
    search/.style={rectangle, draw=black, thick, rounded corners=3pt,
        align=center, minimum width=3.0cm, minimum height=0.62cm,
        fill=orange!12, font=\scriptsize},
    data/.style={cylinder, shape border rotate=90, draw=black, thick,
        align=center, minimum width=2.4cm, minimum height=0.62cm,
        fill=purple!8, font=\scriptsize},
    arr/.style={->, thick, >=Stealth},
    sidelab/.style={text=black!55, font=\tiny\sffamily\itshape, align=center},
]
\node[data]  (lib)   {\texttt{FactorLibrary} $\mathcal{L}$: $N$ entries};
\node[box,   below=of lib]   (dense) {Dense $(N\times M)$ JAX \texttt{int32} array};
\node[box,   below=of dense] (tgt)   {Active poly $q$ as dense vector\\(length $M$)};
\node[search,below=of tgt]   (jax)   {JAX vectorized comparison:\\exact $|$ scalar $|$ permuted};
\node[box,   below=of jax]   (match) {Matched sub-polynomials of $q$};
\node[box,   below=of match] (cands) {\texttt{library\_match} candidates\\(top priority + score bonus)};
\node[box,   below=of cands] (rank)  {\texttt{propose\_splits} ranked output};
\draw[arr] (lib)   -- (dense);
\draw[arr] (dense) -- (tgt);
\draw[arr] (tgt)   -- (jax);
\draw[arr] (jax)   -- (match);
\draw[arr] (match) -- (cands);
\draw[arr] (cands) -- (rank);
\draw[arr, dashed, draw=black!45] (rank.east) -- ++(0.55,0)
    node[sidelab, right, text width=1.0cm] {new\\factorization}
    |- (lib.east);
\end{tikzpicture}
\caption{Top-down \texttt{FactorLibrary}.}
\label{fig:fl-top-down}
\end{subfigure}
\caption{\texttt{FactorLibrary} in the two formulations. }
\label{fig:factor-library-both}
\end{figure*}
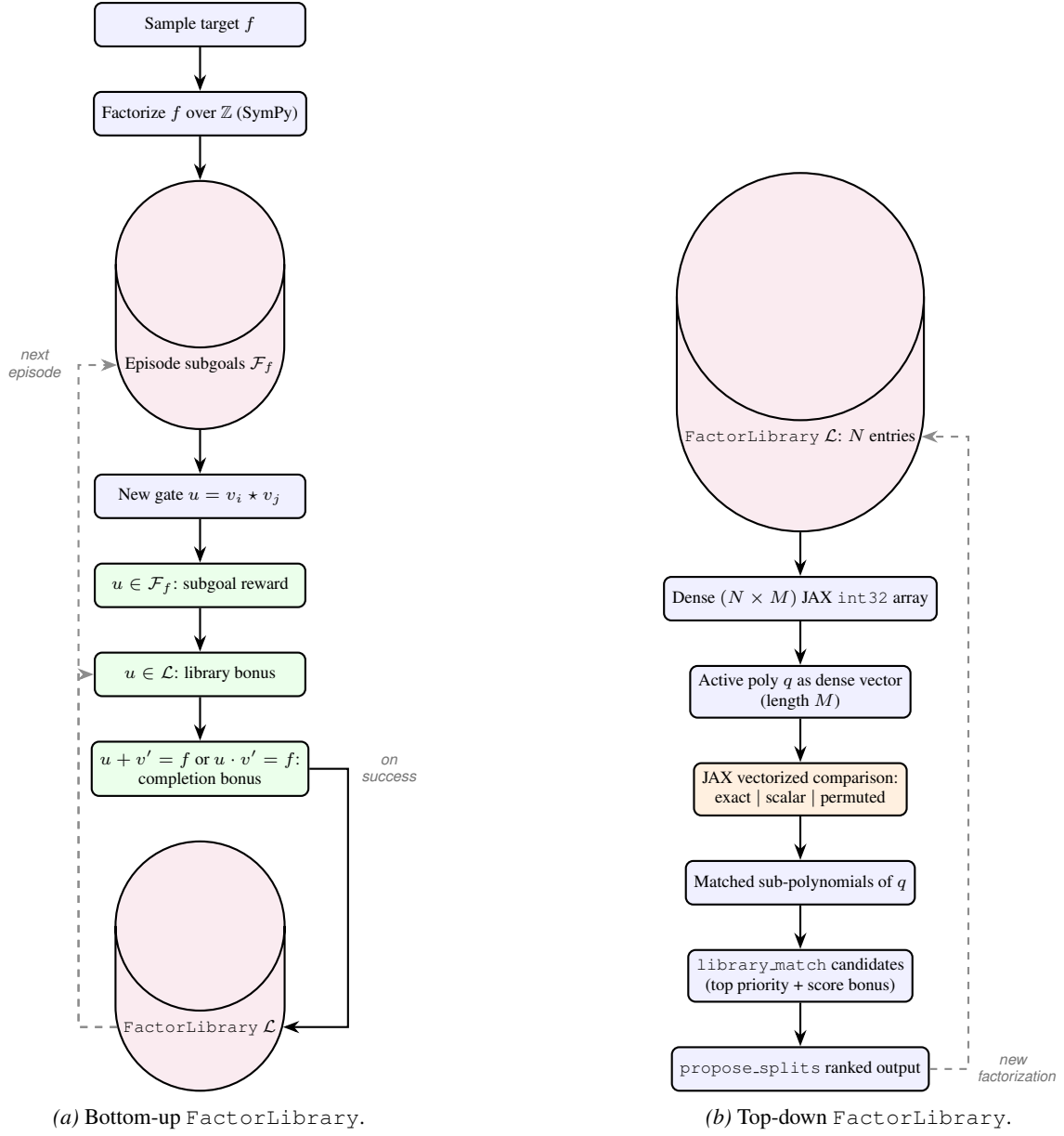

\end{document}